\documentclass{article}
\usepackage{spconf,amsmath,epsfig}
\usepackage{caption}
\usepackage{multirow}
\usepackage{amsfonts,amssymb}
\usepackage{subfigure}
\usepackage{color}

\let\OLDthebibliography\thebibliography
\renewcommand\thebibliography[1]{
  \OLDthebibliography{#1}
  \setlength{\parskip}{0pt}
  \setlength{\itemsep}{0pt plus 0.3ex}
}

\begin{document}\sloppy

\def\x{{\mathbf x}}
\def\L{{\cal L}}

\title{Image Steganography based on Style Transfer}
%
\name{Donghui Hu$^1$, Yu Zhang$^1$, Cong Yu$^1$, Jian Wang$^2$, Yaofei Wang$^2$}

\address{$^1$Hefei University of Technology, $^2$University of Science and Technology of China\\
\{hudh, zaynla\}@hfut.edu.cn, imyucong@163.com, \{wangjian1227, yaofei\}@mail.ustc.edu.cn}

\maketitle

\begin{abstract}
Image steganography is the art and science of using images as cover for covert communications. With the development of neural networks, traditional image steganography is more likely to be detected by deep learning-based steganalysis. To improve upon this, we propose image steganography network based on style transfer, and the embedding of secret messages can be disguised as image stylization. We embed secret information while transforming the content image style. In latent space, the secret information is integrated into the latent representation of the cover image to generate the stego images, which are indistinguishable from normal stylized images. It is an end-to-end unsupervised model without pre-training. Extensive experiments on the benchmark dataset demonstrate the reliability, quality and security of stego images generated by our steganographic network.
\end{abstract}
\section{Introduction}
\label{sec:intro}

Image steganography aims to hide secret information in images without being detected on social networks. The cover image is the one that will be modified to carry the secret message, while the stego image is the output of the steganography algorithm. Traditional image steganography \cite{LSB4607408} mainly adopts the Least Significant Bits (LSB) distribution method to replace the less important pixels in the cover image to carry secret information, however, this method cannot resist steganalysis based on high-dimensional features \cite{LSBsteganalysis1035848,LSBsteganalysis4607422,LSBsteganalysis4607444,LSBsteganalysis5583564,LSBsteganalysis871000}. Then adaptive steganography algorithm \cite{pevny2010using,holub2014universal,holub2012designing} proposed based on the minimum distortion embedding framework, which tends to embed secret messages into textured and noisy regions, but still faces the threat of detection by deep learning-based steganalysis \cite{boroumand2018deep}.


Therefore, many researchers have attempted to combine steganography with deep learning. To date, Generative Adversarial Networks (GANs) \cite{goodfellow2014generative} have been widely used for image generation. It has achieved state-of-the-art performance in many areas, including super-resolution imaging, style transfer, etc. Based on GAN, Hu \textit{et al.} \cite{hu2018novel} and Wang \textit{et al.} \cite{wang2018sstegan} proposed the embedding-free steganography (SwE) method respectively, which learns the steganographic algorithm in an unsupervised manner and directly generates the stego image from the secret message. However, the secret information inevitably interferes with the generated image features and the generated image looks unnatural and thus susceptible to suspicion. Recently, style transfer has received a lot of attention, Wang \textit{et al.} \cite{wang2019stnet} proposed steganography network STNet based on style transfer, but it relies on pre-trained CNNs on ImageNet and introduces additional bias because it is assembled without artistic considerations.



\begin{figure*}[htbp]
\begin{center}
\includegraphics[width=0.8\textwidth]{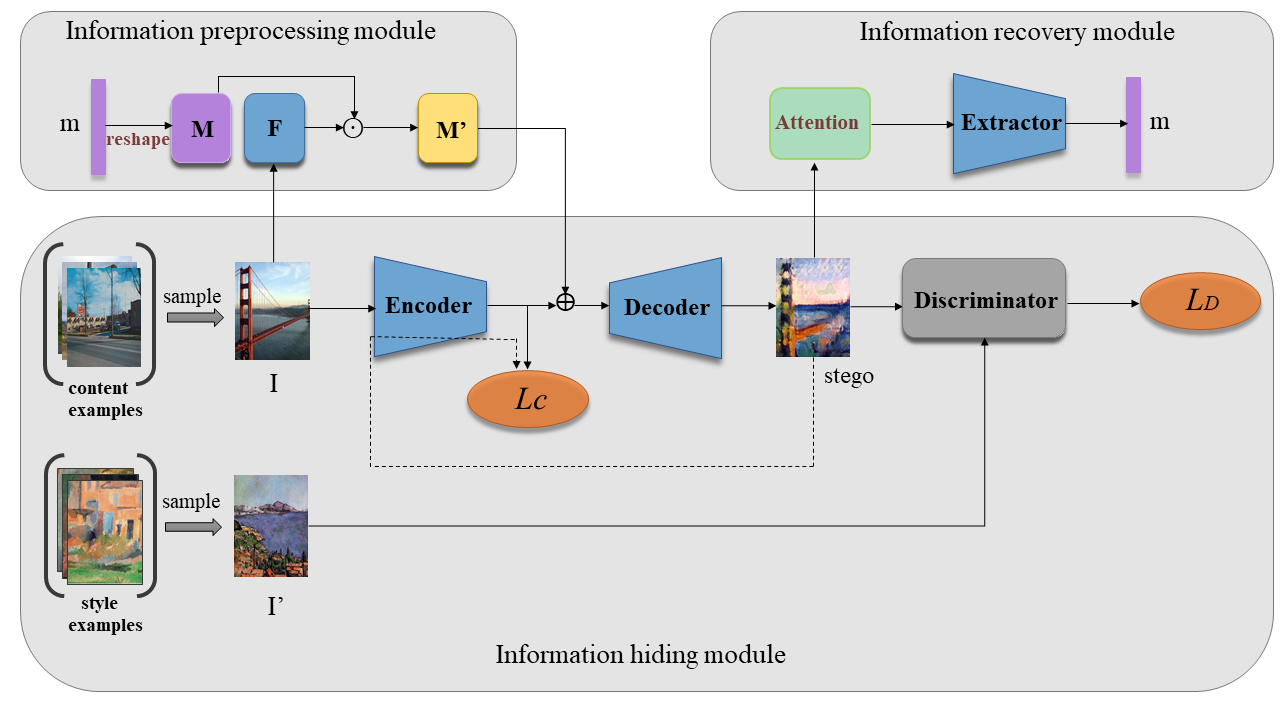}
\end{center}
\caption{The overall framework of the image steganography model based on style transfer.}
\label{1}
\end{figure*}

To resolve the above-mentioned limitation, in this paper, we propose an image steganography model based on style transfer. We adopted an encoder-decoder architecture, which includes information preprocessing module, information hiding module and information recovery module as shown in Fig.\ref{1}. The secret information is input into the information hiding module after passing through the information preprocessing module. The information hiding module includes encoder, decoder and discriminator. The encoder maps the content image into latent representations and the processed secret information is embedded in this process. The decoder generates the stylized stego image. The discriminator is used to distinguish the generated image from the real artistic image to optimize the generated image. The extractor in the information recovery module is used to extract the secret information in the stego image. The steganographic network combined with the style transfer makes it difficult for the steganalyzer to distinguish between the stego image and the stylized cover image.

In summary, the main contributions of this work are as follows:



\begin{itemize}
  \item We designed a steganography model based on style transfer, which embeds secret information while transforming the content image style, thus could reduce the interference of secret information on the generated image. 
  \item It is an end-to-end unsupervised model without pre-training, greatly reducing the computational overhead.
  \item We added channel attention to the extractor and obtained a high message extraction accuracy.
\end{itemize}

\section{Related work}

\label{Sec:related-work}
With the rapid development of social networking and multimedia technologies, image stylization has been widely used in various social media. Style transfer is a method to modify the image style while retaining the original image content. Gatys \textit{et al.} \cite{gatys2015texture,gatys2016image} proposed an iterative optimization method based on a VGG19 network \cite{simonyan2014very}. This method produces high-quality results and is suitable for any inputs, but each optimization step requires forward propagation and backpropagation through the VGG19 network. Sanakoyeu \textit{et al.} \cite{sanakoyeu2018style} proposed a method can be trained on arbitrary unpaired content and style images without back propagation.

Based on style transfer, Yi \textit{et al.} \cite{puyang2018style} proposed a scheme that uses an edge marker embedding scheme to realize information hiding and convert content images into comic images. The comic style generated by this traditional steganography method lacks diversity. Wang \textit{et al.} \cite{wang2019stnet} proposed steganography network STNet based on style transfer, wherein the encoder uses the VGG19 \cite{simonyan2014very} network. However, this model is limited to the combination of a single style image and a content image to show a similar combined image and does not consider the details of the artistic images.

\section{Approach}
\label{Sec:approach}
Our model is shown in Fig.\ref{1}. The steganography framework is mainly divided into three modules, information preprocessing module, information hiding module and information recovery module. The information preprocessing module maps the secret information $m$ to the secret feature $M$, binds $M$ to the content feature $F$ through the dot product operation, and outputs the preprocessed secret feature $M^{'}$. Among them, $F$ is the feature extracted from the content image $I$ by a set of high-pass filters. The information hiding module has an encoder-decoder structure. $I$ is input to the encoder network to generate content features $A$, and then, $A$ is connected with $M^{'}$ to feed to the decoder. The style stego image $stego$ is the output of the decoder. The discriminator is used to measure the gap between $stego$ and the real style image $I^{'}$. We trained an extractor network with channel attention mechanism and used the optimized filters in the first layer of the extractor. The secret information $m$ is recovered from $stego$ by the extractor network. Next, we will introduce each part in detail.

\subsection{Information preprocessing module}
\begin{figure}[htbp]
\begin{center}
\includegraphics[width=0.33\textwidth]{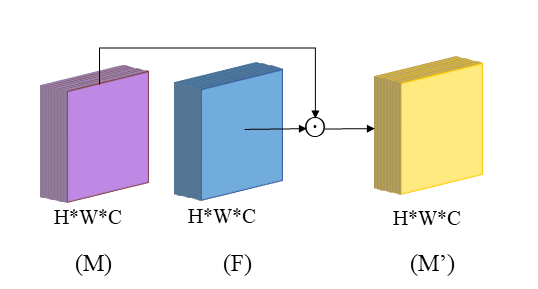}
\end{center}
\caption{Secret message mapping.}
\label{2}
\end{figure} 

To better obtain the content feature $F$ used for secret information mapping, we used a set of high-pass filters (a total of 32 filters, each with a size of 5×5 and three channels), including the high-pass filter of SRM \cite{fridrich2012rich} and its rotating counterparts. The high-pass filter with learnable parameters is used to sample the high-frequency signal from the image. The texture features and spatial position information in the image are extracted as a whole and then passed through a 4-conv layer network and activated by the Leaky ReLU function and the output is the content feature $F$. In this paper, rather than directly concatenate the secret information with the content image that might not fully exploit the semantic information represented by the latent space, as well as corrupting the original features of the input image. Inspired by image enhancement, we propose a new secret information embedding scheme. As shown in Fig. \ref{2}, the combined features:


\begin{eqnarray}
M' &=& M \odot  F,
\label{eq1}
\end{eqnarray}
where $M$ is the secret information matrix, $F$ is the image feature map, $M^{'}\in R^{H \times W \times C}$  is the processed feature by dot product operation. $H$ and $W$ represent the height and width of the matrix, $ C$ is the number of channels.  "$\odot$" represents the dot product operation of the corresponding position elements of the matrix. The secret information matrix is multiplied by the image feature map, which is feature selection. Therefore, the actual embedded secret information is closely related to the texture. Due to the above reasons, when the secret information vector is mapped to the matrix space, it will be forced to learn to arrange the secret information into the matrix according to the image texture characteristics.
\subsection{Information hiding module}

\begin{figure}[htbp]
\begin{center}
\includegraphics[width=0.5\textwidth]{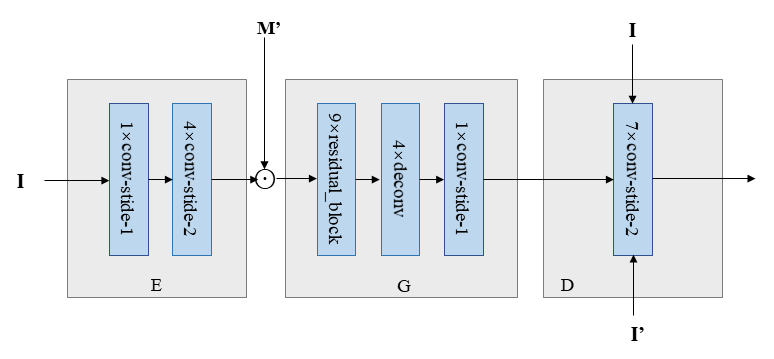}
\end{center}
\caption{Network structure of information hiding module.}
\label{3}
\end{figure}

Information hiding module is designed on the basis of the style transfer model \cite{sanakoyeu2018style}. The model mainly includes encoder $E$, decoder $G$ and discriminator $D$. As shown in Fig. \ref{3}, the encoder network consists of 5 convolutional layers: 1$\times$conv-stride-1 and 4$\times$conv-stride-2. The decoder network contains 9 residual blocks \cite{he2016deep}, 4 up-sampling blocks and 1$\times$conv-stride-1. The discriminator is a fully convolutional network with 7$\times$conv-stride-2 layers.

The style of the generated stego image is controlled by $D$. Therefore, the style loss function is also the discriminator loss function to distinguish the output stylized image $G(E(x_{i}))$ from the real style sample $s_{i} \in I^{'}$ , $x_{i} \in I$ is a content image. The style loss function of our model is given by 

\begin{eqnarray}
\begin{aligned}
L_{style} = &\mathop{\mathbb{E}}_{s \sim p_{style}}[\log D(s)]+ \\
&\mathop{\mathbb{E}}_{x \sim p_{x}(x)}[\log (1-D(G(E(x))))].
\end{aligned}
\label{eq2}
\end{eqnarray}


We map the secret information into a feature map and hide the secret while generating the stylized image.
The full objective function of our model is 

\begin{eqnarray}
\begin{aligned}
L(E,G,D) &=& \lambda L_{style}+L_{c} (E,G),
\end{aligned}
\label{eq3}
\end{eqnarray}
$E$ and $G$ are trained on the content image set and the style image set to generate stylized stego images. $\lambda$ controls the relative importance of style loss.

\subsection{Information recovery module}


When recovering secret information from the received stego image, a positive number represents $1$ and a negative number represents $0$. The secret loss function is used to measure the difference between the output of the extractor network and the original secret information and is given by

\begin{eqnarray}
L_{m} \ = \ min(max(0, (m-\widehat{m})^{2}-\delta))\nonumber \\
m\sim p_{secret}(m),
\label{eq4}
\end{eqnarray}
where $m$ is the secret information, $\widehat{m}$ is the extracted secret information by the extractor network, $\delta$ is the upper threshold of the error accuracy of the secret information.
We use a set of high-pass filter in the first layer of the extractor to better obtain image features with secret information and improve the secret information recovery capability of the extractor network. In the extractor of our model, we also added a channel attention mechanism \cite{hu2018squeeze} so that the extractor can pay attention to image features with rich textures because the secret information is embedded in this area. We first extract the feature map $I_{f_{c}}$ from the stego image. Then the $I_{f_{c}}$ will be fed into the attention model and the model is formulated as

\begin{eqnarray}
s_{c} &=& \delta(W_{2}(\delta(W_{1}(\phi(I_{f_{c}}))))),
\label{eq6}
\end{eqnarray}
where $c$ represents the number of feature channels, $\delta$ is the ReLU \cite{nair2010rectified} function, $\phi$ is the pooling layer. $W_{1}\in R^{\frac{c}{r}\times c}$, $W_{2}\in R^{ c \times \frac{c}{r}}$ the hyperparameter $r$ is reduced to a coefficient. Finally, multiply the learned activation value $s_{c}$ by the original feature $I_{f_{c}}$ to obtain a feature map with different channel attention, 

\begin{equation}
\begin{aligned}
\widetilde{x_{c}} &=& s_{c} \cdot I_{f_{c}}.
\end{aligned}
\label{eq10}
\end{equation}
Channel attention mechanism make the extractor focus on regions with rich texture and improve the extraction of secret information.

\section{Experiment}
\label{Sec:experiment}
In the training process, we use the Places365 \cite{zhou2014learning} dataset as the content image, and the WikiArt \cite{karayev2013recognizing} dataset style image block is sampled. The dataset Places$365$ contains approximately $1.8$ million images from $365$ scene categories, each with a maximum of $5000$ images. Therefore, it can provide rich image content for online learning. The WikiArt art image data set contains a large number of art style images of various genres and styles, which are used to provide a variety of training styles, such as Cézanne, Monet, Gauguin and other styles of artistic images. During the training process, we randomly crop an area with a size of $256\times256$. We use batch size $1$, learning rate $0.0002$ and Adam \cite{kingma2014adam} optimizer training for $300,000$ iterations. A string of $0$ or $1$ bits is randomly generated as the secret information vector. In information recovery module, the recovered information is compared with the input secret information in bit unit, which is used as an evaluation index for the accuracy of information extraction.

\subsection{Reliability}


  

Here we evaluate the accuracy of secret information recovery. To improve the recovery rate of secret information, we optimized the filters for extracting the texture features of the stego image and added a channel attention mechanism to the extractor. To verify whether the above two modules are useful for message extraction, we designed three ablation experiments at a message embedding size of 1000 bits. The experimental results are shown in Table \ref{table6}. As we can see that with the optimized filters and channel attention can improve the recovery accuracy of secret information. 

Then we compare the accuracy with STNet \cite{wang2019stnet}, which is also a steganography network based on style transfer, at different embedding size. The experimental results are shown in Table \ref{Comparison of message recovery accuracy}. We can see that the message recovery performance of these two methods is comparable, although our method is slightly weaker, but in practice we can also achieve $100\%$ after using the error correction code. In addition, STNet is limited to a blunt combination of images and styles, and the quality of the generated images is weaker than our method as shown below.


\begin{table}[h]
  \caption{Message recovery accuracy under different experimental conditions.}
  \centering
  \label{table6}
\begin{tabular}{c|ccc}
\hline
Channel attention & w/ & w/o & w/ \\ \hline
Optimized filters & w/o & w/ & w/ \\ \hline
Recovery accuracy & 0.96 & 0.98 & 0.99 \\ \hline
\end{tabular}
\end{table}

\begin{table}[h]
  \caption{Comparison of message recovery accuracy of different methods at different embedding size.}
  \centering
  \label{Comparison of message recovery accuracy}
\begin{tabular}{c|cc}
\hline
Embedding size (bits) & 1000 & 2000  \\ \hline
STNet \cite{wang2019stnet}  & 0.99 & 0.99  \\ \hline
Ours & 0.99 & 0.98  \\ \hline
\end{tabular}
\end{table}

\subsection{Quality}
\begin{figure}[h]
\begin{center}
\includegraphics[width=0.45\textwidth]{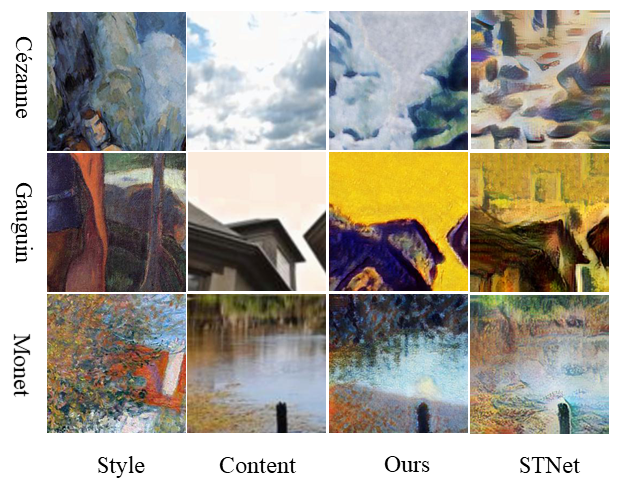}
\end{center}
\caption{Comparison of the images generated by our steganography model with the STNet model.}
\label{fig:res}
\end{figure}

In Fig.\ref{fig:res}, we show the stylized stego image generated by our model and STNet. Stego images generated by our steganography model are like oil paintings created by artists, which are highly artistic. In order to quantitatively evaluate the quality of the stylized stego images generated by our model and STNet model, we evaluate these images through SSIM. The results are displayed in Table \ref{table5}. In all three art styles, the SSIM evaluation index of images generated by our model is comparable to that of STNet. In addition,STNet can only be trained to combine style images and content images and cannot learn the general features of one class of style images.In contrast, the image generated by our method is more like an artistic oil painting, and our method is more artistic.

\begin{table}[h]
  \caption{The quality of the stego images.}
  \centering
  \label{table5}
  \scalebox{1.0}{
  
  \setlength{\tabcolsep}{3mm}
  \begin{tabular}{c|ccc}
  \hline
  SSIM &Cézanne &Gauguin &Monet \\
  \hline
 	    STNet& 0.93 &0.92 &0.92\\\hline
		Ours&0.93 &0.93 &0.91\\\hline
  \end{tabular}}
  \end{table}


\subsection{Security}
\begin{figure}[h]
\begin{center}
\includegraphics[width=0.45\textwidth]{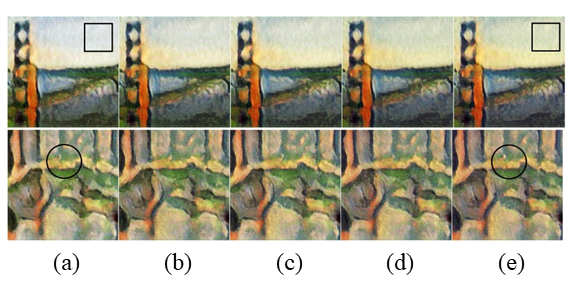}
\end{center}
\caption{The training step size has a subtle effect on the generated image. The color difference can be seen at the indicated position.}
\label{steps}
\end{figure}

Unlike the security pursued by traditional steganography, the goal is to make stego and cover indistinguishable. Our model aims to make the generated stylized steganographic images indistinguishable from the normal stylized images. Moreover, our model generates subtly different images for different training under the same message embedding, making the steganalysis network more difficult to train. Fig. \ref{steps} shows the stylized stego images generated by the same model in different iterations. The images from (a) to (e) have slight color differences in the same style. This is consistent with a realistic scenario, as the images created are not the same each time, even for the same work created by the same artist. From a security point of view, generating similar but different steganographic images from the same model can further resist the training of steganalysis networks.

Our model embeds secret information during the stylization of content images, and current steganalysis cannot determine whether the modified part of the generated images come from style transfer or information embedding. And it is meaningless for steganalyzer to distinguish between the generated image and the content image, because the generated stego image is based on the stylized image. In our security experiment design, after receiving the generated image, the steganalyzer needs to determine whether the stylized image contains secret information. Therefore the cover image is a stylized image without secret information, stego image is a stylized image with secret information. We use SRNet \cite{boroumand2018deep}, the typical steganalysis network, to evaluate the security of our model. The experimental data settings are: the artistic style of the generated image is Cézanne, the style image generated by the style transfer model \cite{sanakoyeu2018style} is cover, and the style stego image generated by our model is stego. 

To evaluate the security of our stylized stego images, we distinguish three possible kinds of steganalyzers: 1) An Ignorant Steganalyzer who has obtained our model, but does not know which iterative model and only has one iterative model. Therefore, we set the training dataset as the cover and stego image pairs generated by corresponding models saved in a certain iteration, the testing dataset as the image pair generated by corresponding models saved in another iteration; 2) A Knowledgeable Steganalyzer who knows different iterative models, but the number of iterations of the generated stego is not known; and 3) An Omniscient Steganalyzer who knows iterations of generated stego. The experiment results are shown in Table \ref{table1}. Under the embedding size of 1000 bits, for the ignorant steganalyzer, the detection accuracy of ours is 0.55, $20\%$ less than STNet, which is equivalent to blind guessing. The knowledgeable steganalyzer achieves a detection accuracy of 0.60 and omniscient steganalyzer achieves a detection accuracy of 0.71, $27\%$ and $22\%$ less than STNet, respectively. These results show that our security is stronger than STNet in different situations and difficult to be detected by ignorant steganalyzer in practical applications.




  
 \begin{table}[t]
  \caption{Classification accuracy of SRNet under different steganalyzers.}
  \centering
  \label{table1}
\begin{tabular}{c|c|cc}
\hline
\multicolumn{2}{c|}{Steganalyzer\textbackslash{}Embedding size (bits)} & 1000 & 2000  \\ \hline
\multirow{2}{*}{Ignorant}                      & STNet                & 0.83 & 0.98   \\ \cline{2-4}
                                               & Ours                 & \textbf{0.55} & \textbf{0.59}           \\ \hline
\multirow{2}{*}{Knowledgeable}                 & STNet                & 0.87 & 0.99           \\ \cline{2-4}
                                               & Ours                 & \textbf{0.60} & \textbf{0.73}           \\ \hline
\multirow{2}{*}{Omniscient}                     & STNet                & 0.93 & 0.99  \\ \cline{2-4}
                                               & Ours                 & \textbf{0.71} & \textbf{0.84}           \\ \hline
\end{tabular}
\end{table}


\section{Conclusions}
\label{Sec:conclusions}
In this paper, we propose a steganography model based on style transfer, which hides secret information during the process of style transfer. The stylized stego images generated by this model are artistic in nature, making it less susceptible to suspicion when transmitted over the Internet. In the experiments, we test the performance of the proposed method compared with the STNet. Our method has a high message recovery accuracy and is more practical and is more artistic, because it can capture the subtle details of style images through training. In terms of security, in resisting deep-learning based steganalysis network detection, our method has high security performance in all three scenarios, satisfying the pursuit of steganography. Our future work is to improve the robustness and embedding rate of the method.


\bibliographystyle{IEEEbib}
\bibliography{icme2022template}

\end{document}